\documentclass[11pt,a4paper]{article}
\usepackage[hyperref]{emnlp-ijcnlp-2019}
\usepackage{times}
\usepackage{latexsym}

\usepackage{url}

\usepackage{url}
\usepackage{graphicx}  
\usepackage{subfigure}
\usepackage{amsmath}
\usepackage{multirow}
\usepackage{subfigure}
\usepackage{arydshln}
\usepackage{color}
\usepackage{amsmath,amssymb,bm}
\usepackage{algorithm}
\usepackage{algorithmic}
\usepackage[american]{babel}
\usepackage[c]{esvect}
\usepackage{footnote}
\usepackage{enumerate}
\usepackage{CJKutf8}
\usepackage{footnote}
\usepackage{mathtools}
\makeatletter
\newcommand\argmax{\mathop{\operator@font argmax}}
\newcommand\argmin{\mathop{\operator@font argmin}}
\makeatother

\DeclarePairedDelimiter\ceil{\lceil}{\rceil}

\aclfinalcopy 

\title{Cross-Lingual Dependency Parsing Using Code-Mixed TreeBank}

\author{Meishan Zhang$^1$ ~~\textnormal{and}~~ Yue Zhang$^2$\thanks{~~Corresponding author.} ~~\textnormal{and}~~ Guohong Fu$^3$ \\
1. School of New Media and Communication, Tianjin University, China \\
2. Institute of Advanced Technology, Westlake Institute for Advanced Study \\
3. Institute of Artificial Intelligence, Soochow University, China \\
{\tt mason.zms@gmail.com,} \\
{\tt zhangyue@westlake.edu.cn,} \\
{\tt ghfu@hotmail.com }
}

\date{}

\begin{document}
\maketitle
\begin{abstract}
Treebank translation is a promising method for cross-lingual transfer of syntactic dependency knowledge.
The basic idea is to map dependency arcs from a source treebank to its target translation according to word alignments.
This method, however, can suffer from imperfect alignment between source and target words.
To address this problem, we investigate syntactic transfer by code mixing, translating only confident words in a source treebank.
Cross-lingual word embeddings are leveraged for transferring syntactic knowledge to the target from the resulting code-mixed treebank.
Experiments on University Dependency Treebanks show that code-mixed treebanks are more effective than translated treebanks,
giving highly competitive performances among cross-lingual parsing methods.
\end{abstract}

\section{Introduction}
Treebank translation \cite{tiedemann2014treebank,tiedemann2015improving,tiedemann2016synthetic} has been considered
as a method for cross-lingual syntactic transfer.
Take dependency grammar for instance.
Given a source treebank, machine translation is used to find target translations of its sentences.
Then word alignment is used to find mappings between source and target words,
so that source syntactic dependencies can be projected to the target translations.
Following, a post-processing step is applied by removing unaligned target words,
in order to ensure that the resulting target syntax forms a valid dependency tree,
The method has shown promising performance for unsupervised cross-lingual dependency parsing
among transfer methods \cite{mcdonald2011multi,tackstrom2012cross,rasooli2015density,guo2016representation}.

\begin{figure}[tb]
	\begin{center}
		\subfigure{ \label{fig:dependency}
			\centering{\includegraphics[scale=0.75]{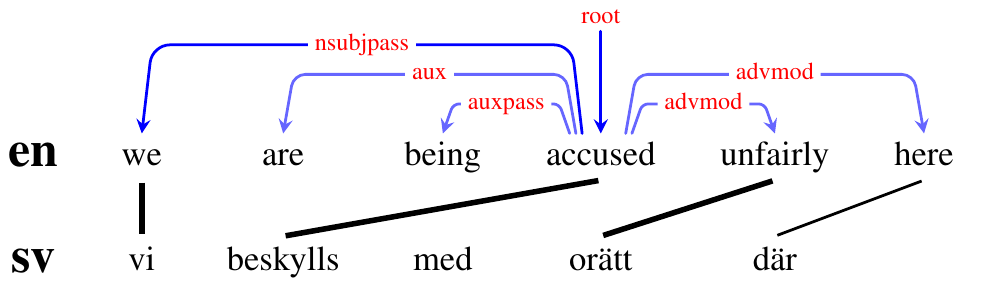}}			
		}
		\setcounter{subfigure}{0}
		\subfigure[full-scale translation.]{\label{fig:baseline}
			\centering{\includegraphics[scale=0.75]{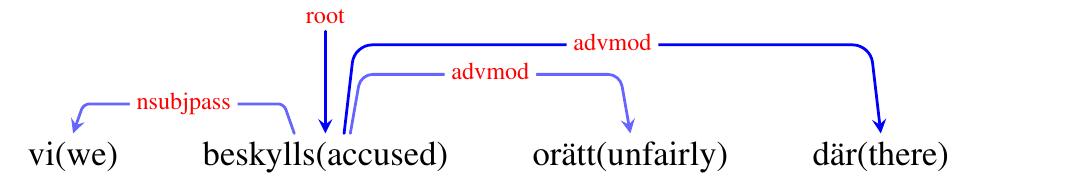}}
		}
		\subfigure[this method, partial translation.]{\label{fig:parser}
			\centering{\includegraphics[scale=0.75]{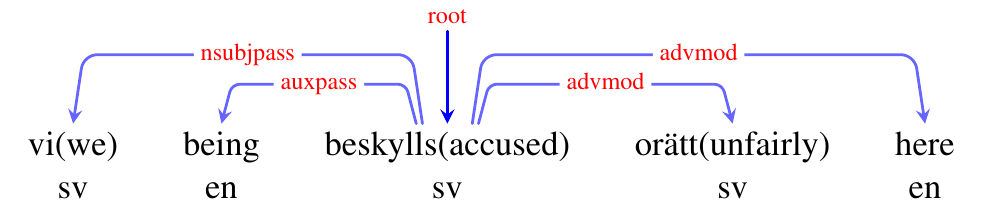}}
		}
		\caption{An example to illustrate our method, where the source and target languages are English (en) and Swedish (sv), respectively.}\label{fig:overall:method}
	\end{center}
\end{figure}

The treebank translation method, however, suffers from various sources of noise.
For example, machine translation errors directly affect the resulting treebank,
by introducing ungrammatical word sequences.
In addition, the alignments between source and target words may not be isomorphic
due to inherent differences between languages or paraphrasing during translation.
For example, in the case of Figure \ref{fig:overall:method},
the English words ``are'' and ``being'',
and the Swedish word ``med'', do not have corresponding word-level translation.
In addition, it can be perfect to express ``as soon as they can'' using ``very quickly'' in a translation,
which looses word alignment information because of the longer span.
Finally, errors in automatic word alignments can also bring noise.
Such alignment errors can directly affect grammaticality of the resulting target treebank
due to deletion of unaligned words during post-processing,
or cause lost or mistaken dependency arcs.

We consider a different approach for translation-based syntactic knowledge transfer,
which aims at making the best use of source syntax with the minimum noise being introduced.
To this end, we leverage recent advances in cross-lingual word representations,
such as cross-lingual word clusters \cite{tackstrom2012cross} and cross-lingual word embeddings \cite{guo2015cross},
which allow words from different languages to reside within a consistent feature vector space
according to structural similarities between words.
Thus, they offer a bridge on the lexical level between different languages \cite{ammar2016many}.

A cross-lingual model can be trained by directly using cross-lingual word representations on a source treebank \cite{guo2015cross}.
Using this method, knowledge transfer can be achieved on the level of token correspondences.
We take this approach as a naive baseline.
To further introduce structural feature transfer, we transform a source treebank
into a code-mixed treebank by considering word alignments between a source sentence
and its machine translation.
In particular, source words with highly confident target alignments are translated into target words
by consulting the machine translation output,
so that target word vectors can be directly used for learning target syntax.
In addition, continuous spans of target words of the code-mixed treebank are reordered according to the translation,
so that target grammaticality can be exploited to the maximum potent.

We conduct experiments on Universal Dependency Treebanks (v2.0) \cite{mcdonald2013universal,nivre2016universal}.
The results show that a code-mixed treebank can bring
significantly better performance compared to a fully translated treebank,
resulting in averaged improvements of 4.30 points on LAS.
The code and related data will be released publicly available  under Apache License 2.0.\footnote{https://github.com/zhangmeishan/CodeMixedTreebank}

\section{Related Work}
Existing work on cross-lingual transfer can be classified into two categories.
The first aims to train a dependency parsing model on source treebanks \cite{mcdonald2011multi,guo2016distributed,guo2016representation},
or their adapted versions \cite{zhao2009cross,tiedemann2014treebank,hongmin:2017:singapore} in the target language.
The second category, namely annotation projection, aims to produce a set of large-scale training instances
full of automatic dependencies
by parsing parallel sentences \cite{hwa2005bootstrapping,rasooli2015density}.
The two broad methods are orthogonal to each other,
and can both make use of the lexicalized dependency models trained with cross-lingual word representations \cite{rasooli2017cross,rasooli2019low}.

{\bf Source Treebank Adaption.}
There has been much work on unsupervised cross-lingual dependency parsing by direct source treebank transferring.
Several researchers investigate delexicalized models where only non-lexical features are used in the models \cite{zeman2008cross,cohen2011unsupervised,mcdonald2011multi,naseem2012selective,tackstrom2013target,rosa2015klcpos3}.
All the features in these models are language independent,
and are consistent across languages and treebanks.
Thus they can be applied into target languages directly.

Subsequent research proposes to exploit lexicalized features to enhance the parsing models,
by resorting to cross-lingual word representations \cite{tackstrom2012cross,guo2015cross,duong2015low,duong2015cross,zhang2015hierarchical,guo2016representation,ammar2016many,wick2016minimally,de2018parameter}.
Cross-lingual word clusters and
cross-lingual word embeddings are two main sources of features for transferring knowledge
between source and target language sentences.
These studies enable us to train lexicalized models on code-mixed treebanks as well.
Thus here we integrate the cross-lingual word representations as well,
which gives more direct interaction between source and target words.

Our work follows another mainstream method of this line of work,
namely treebank translation \cite{tiedemann2014treebank,tiedemann2015improving,tiedemann2016synthetic},
which aims to adapt an annotated source treebank into the target language by machine translation.
In addition, the target-side sentences are produced by machine translation.
Previous work aims to build a well-formed tree \cite{tiedemann2016synthetic} from source dependencies,
solving word alignment conflicts by heuristic rules.
In contrast, we use partial translation instead to avoid unnecessary noise.

{\bf Annotation Projection.}
The annotation projection approach relies on a set of parallel sentences between the source and target languages \cite{hwa2005bootstrapping,ganchev2009dependency}.
In particular, a source parser trained on the source treebank
is used to parse the source-side sentences of the parallel corpus.
The source dependencies are then projected onto the target sentences according to word alignments.
Different strategies can be applied for the dependency projection task \cite{ma2014unsupervised,rasooli2015density,xiao2015annotation,agic2016multilingual,schlichtkrull2017cross}.
For example, one can project only dependency arcs whose words
are aligned to target-side words with high confidence \cite{lacroix2016frustratingly}.
The resulting treebank can be highly noisy due to the auto-parsed source dependency trees.
Recently \newcite{lacroix2016frustratingly} and \newcite{rasooli2017cross} propose to filter the results
from the large-scale parallel corpus.
Our work is different in that the source dependencies are from gold-standard treebanks.

\section{Dependency Parsing}
\begin{figure}[tb]
\begin{center}
\includegraphics[scale=0.9]{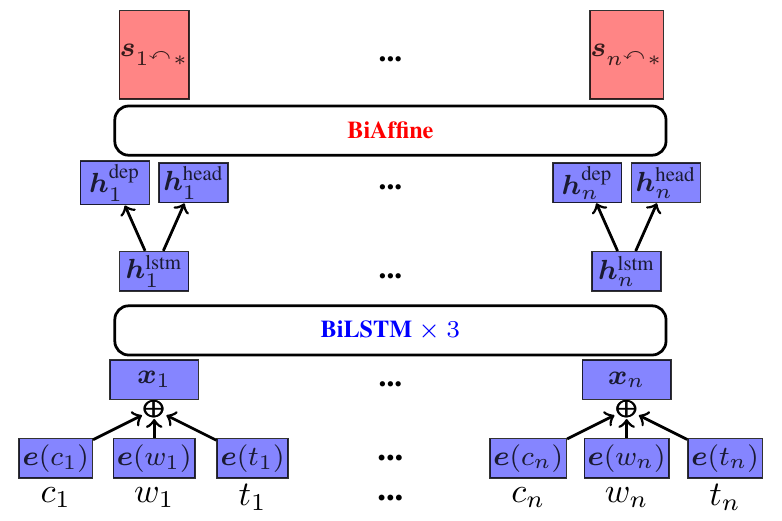}
\caption{The overall architecture of the BiAffine parser. }\label{biafine}
\end{center}
\end{figure}

We adopt a state-of-the-art neural BiAffine parser \cite{dozat2016deep} as the baseline,
which has achieved competitive performances for dependency parsing.
The overall architecture is shown in Figure \ref{biafine}.
Given an input sentence $w_1\cdots w_n$,
the model finds its embedding representations of each word $\bm{x}_1\cdots\bm{x}_n$, where
\begin{equation}
 \bm{x}_i = \bm{e}(w_i) \oplus \bm{e}(c_i) \oplus \bm{e}(t_i)
\end{equation}
Here $c_i$ denotes the word cluster of $w_i$, and $t_i$ denotes the POS tag.
We exploit cross-lingual word embeddings,  clusters and also universal POS tags, respectively,
which are consistent across language and treebanks.

A three-layer deep bidirectional long short term memory (LSTM) neural structure
is applied on $\bm{x}_1\cdots\bm{x}_n$ to obtain hidden vectors $\bm{h}_1^{\text{LSTM}}\cdots\bm{h}_n^{\text{LSTM}}$.
For head finding, two nonlinear feed-forward neural layers are used on $\bm{h}_1^{\text{LSTM}}\cdots\bm{h}_n^{\text{LSTM}}$ to
obtain $\bm{h}_1^{\text{dep}}\cdots\bm{h}_n^{\text{dep}}$ and
$\bm{h}_1^{\text{head}}\cdots\bm{h}_n^{\text{head}}$.
We compute the score for each dependency $i^{\overset{\text{}}{\curvearrowleft}}j$ by:
\begin{equation}
 \bm{s}_{i^{\overset{\text{}}{\curvearrowleft}}j} = \text{BiAffine}(\bm{h}_i^{\text{dep}}, \bm{h}_j^{\text{head}})
\end{equation}
The above process is also used for scoring a labeled dependency $i^{\overset{\text{l}}{\curvearrowleft}}j$,
by extending the 1-dim vector $\bm{s}$ into $L$ dims, where $L$ is the total number of dependency labels.

When all scores are ready, a softmax function is used at each position for all candidate heads and labels,
and then the normalized scores are used for decoding and training.
For decoding, we exploit the MST algorithm to ensure tree-structural outputs.
For training, we accumulate the cross-entropy loss at the word-level by treating the normalized scores as prediction probabilities.
The reader is referred to \newcite{dozat2016deep} for more details.

\section{Code-Mixed Treebank Translation}
We derive code-mixed trees from source dependency trees by partial translation,
projecting words and the corresponding dependencies having high-confidence alignments with machine-translated target sentences.
Our approach assumes that sentence level translations and alignment probabilities are available.
The motivation is to reduce noise induced by problematic word alignments.

We adopt the word-level alignment strategy,
which has been demonstrated as effective as phrase-level alignment yet much simpler \cite{tiedemann2014treebank,tiedemann2015improving,tiedemann2016synthetic}.
Give a source sentence $e_1 \cdots e_n$ and its target language translation $f_1 \cdots f_m$,
$p(e_i|f_j)$  denotes  the probability of word $f_j$ being aligned with $e_i$ ($0 \leq i \leq n $ and $ 0 < j \leq m$),
where $e_0$ denotes a null word, indicating the no alignment probability for one target word.

The translation process can be conducted by three steps:
\begin{itemize}
  \item[(1)] word substitution, which incrementally substitutes the source words with the target translations;
  \item[(2)] word deletion, which removes several unaligned source words;
  \item[(3)] sentence reordering, which reorders the partially translated sentence,
ensuring local target language word order.
\end{itemize}
Algorithm \ref{alg:algorithm} shows pseudocode for code-mixing tree translation,
where line 1-8 denotes the first step,
line 9-16 denotes the second and line 17 denotes the last step.

\begin{algorithm}[tb]
\caption{The process of tree translation.}
\label{alg:algorithm}
\textbf{Input}: ~~$e_1\cdots e_n$,~~$f_1\cdots f_m$,~~$\text{T}_e$, ~~$\{ p(e|f) \}$, ~~$\lambda$ \\
\textbf{Output}: ~~$\text{T}_{\text{mix}}$
\begin{algorithmic}[1] 
\STATE Let $\text{T}_{\text{mix}} = \text{T}_e$.
\FOR {$j \in [1, m]$ }
\STATE $a_j, p_j \leftarrow \argmax_{i} p(e_i|f_j), \max_{i} p(e_i|f_j)$
\ENDFOR
\STATE $S_{\text{sub}}$ $\leftarrow$  \textsc{Select}($\{\langle f_j, a_j, \bm{v}\text{=}p_j \rangle\}, \lambda$)
\FOR {$f_j, a_j \in S_{\text{sub}}$}
\STATE $\text{T}_{\text{mix}}$ $\leftarrow$ \textsc{Substitute}($f_j, a_j,\text{T}_{\text{mix}}$)
\ENDFOR
\STATE $D = [1, n] - \{ a_j\}|_{j\in[1,m]}$
\FOR {$i \in D$}
\STATE $r_i = \sum_{j} p(e_i|f_j)$
\ENDFOR
\STATE $S_{\text{del}}$ $\leftarrow$  \textsc{Select}($\{\langle e_i, \bm{v}\text{=-}r_i \rangle\}, \lambda$)
\FOR {$e_i \in S_{\text{del}}$}
\STATE $\text{T}_{\text{mix}}$ $\leftarrow$ \textsc{Delete}($e_i, \text{T}_{\text{mix}}$)
\ENDFOR
\STATE $\text{T}_{\text{mix}}$ $\leftarrow$ \textsc{Reorder}($\text{T}_{\text{mix}}, f_1\cdots f_m$)
\end{algorithmic}
\end{algorithm}

There is a hyper-parameter $\lambda$ to control the overall ratio of translation.
The function \textsc{Select}($ S, \lambda$) is to obtain a subset of $S$ by ratio $\lambda$
with top element values as indicated by $\bm{v}$ inside $S$.
If $\lambda = 0$, it is still the source language dependence tree, since no source word is substituted or deleted.
In this condition, our method is equal to \newcite{guo2015cross} by bridging source and target dependency parsing with universal word representations.
If $\lambda = 1$, the resulting tree is a fully translated target dependency tree, as all words are target language produced by translation.
In this setting, our method is equal to \newcite{tiedemann2016synthetic} where the only difference the our baseline parsing model.
Thus our method can be regarded as a generalization of both source-side training \cite{guo2015cross}
and fully translated target training \cite{tiedemann2016synthetic}
with fine-grained control over translation confidence.

\begin{figure}[tb]
	\begin{center}
		\subfigure[one-to-one.]{\label{fig:one:one}
			\centering{\includegraphics[scale=0.75]{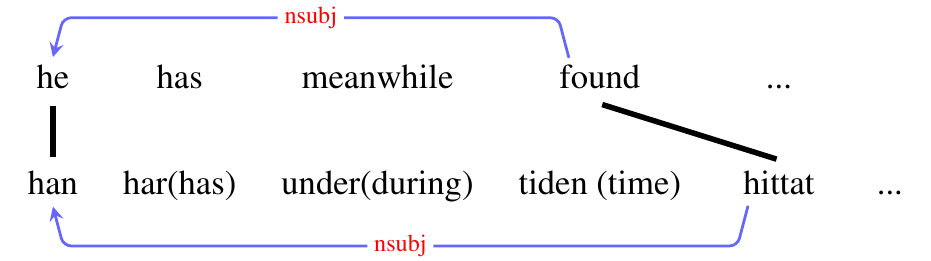}}
		}
		\subfigure[many-to-one.]{\label{fig:many:one}
			\centering{\includegraphics[scale=0.75]{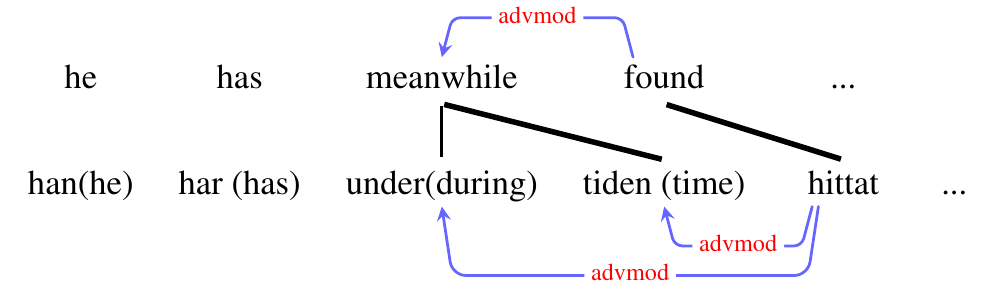}}
		}
		\caption{Examples of word substitution, where the thicker lines indicates more confident alignments.}\label{fig:substitution}
	\end{center}
\end{figure}

\subsection{Word Substitution}
Word substitution is the key step for producing a target treebank.
We first obtain the most confidently aligned source word $e_{a_j}$ for each target word $f_j$ as well as
their alignment probability $p_j = p(e_{a_j}|f_j)$, as shown by line 3 in Algorithm \ref{alg:algorithm}.
Then we sort the target words by these alignment probabilities,
choosing the top $\ceil*{m\lambda}$ words with highest alignment probabilities for substitution.
The sorting and choosing is reflected in line line 5 of Algorithm \ref{alg:algorithm}.
Finally for each chosen word $f_j$ and its aligned source word $e_{a_j}$ ,
we replace the source word $e_{a_j}$ by $f_j$, as shown by line 7 in Algorithm \ref{alg:algorithm}.

One key of the substitution is to maintain the corresponding dependency structures.
If $e_{a_j}$ and $f_j$ bares a one-one mapping,
with no other target word being aligned with $e_{a_j}$,
the source dependencies are kept unchanged, as shown by Figure \ref{fig:one:one}.
While if two or more words (i.e., $f_{j_1},\cdots, f_{j_k} (j_1 < \cdots < j_k)$) are aligned
to $e_{a_j}$,
we simply link all words to $e_{a_j}$, with the same dependency label as the original dependency arc.
Figure \ref{fig:many:one} illustrates this condition.
Both the Swedish words ``under'' and ``tiden'' are headed to ``hittat'' (the Swedish translation of English word ``found'')
by the dependency label ``advmod'' inherited from the source English side .
Note that the POS tags of the substituted words are the same as the corresponding source words.

\subsection{Word Deletion}
There can be source words to which no target word is aligned.
These words are typically functional words belonging to source language only,
such as ``the'', ``are'' and ``have''.
We remove such words to produce dependency trees
that are close in syntax to the target language.

In particular,
we accumulate the probabilities of $p(e_i|f_j)$ for the source word $e_i$ who has no aligned target word:
\[ r_i = \sum_{j} p(e_i|f_j)  \text{~~~~ (line 11 in Algorithm \ref{alg:algorithm})}, \]
where we traverse all target words to sum their alignment probabilities with $e_i$.
The value of $r_i$ can be interpreted as the confidence score of retention.
The words with lower retention scores should be more preferred to be deleted,
as these words have lower probabilities aligning with some word of the target language sentence.
Concretely, we collect all source words with no aligned target words,
computing their retention values,
and the selecting a subset of these words with the lowest retention values
by the hyper-parameter $\lambda$ (line 13 in Algorithm \ref{alg:algorithm}).
Finally we delete all the selected words (line 15 in Algorithm \ref{alg:algorithm}).

\begin{figure}[tb]
\begin{center}
\includegraphics[scale=0.75]{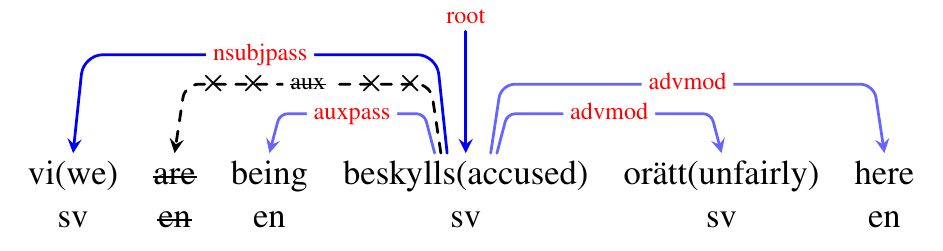}
\caption{An example of word deletion. }\label{worddel}
\end{center}
\end{figure}

Figure \ref{worddel} shows an example of word deletion.
The two words ``are'' and ``being'' both have no aligned words in the other side,
and meanwhile ``are'' has a lower retention score compared with ``being''.\footnote{Both
the two words are only related to the Swedish word ``med'', but $p(\text{are}|\text{med})$  is slightly lower.}
Thus the source word ``are'' is prefer to be deleted.
In most cases, the deleted words are leaf nodes,
which can be unattached to the resulted dependency tree and deleted them directly.
In case of exceptions,
we simply reset the corresponding heads of its child nodes by the head of $e_i$ (i.e., $\text{h}_{e_i}$) instead.
For example, a dependency  $w_i^{\overset{\text{}}{\curvearrowleft}}e_i$
is changed into $w_i^{\overset{\text{}}{\curvearrowleft}}\text{h}_{e_i}$.

\begin{figure}[tb]
\begin{center}
	\subfigure[full sentence]{\label{fig:reorder1}
		\centering{\includegraphics[scale=0.75]{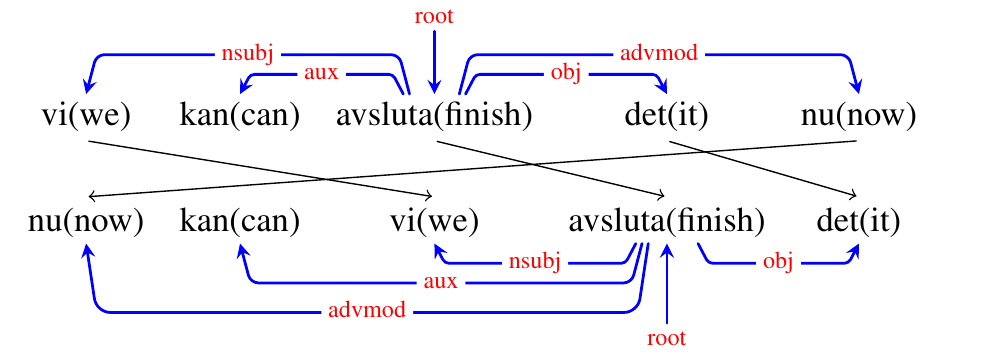}}
	}
	\subfigure[two spans]{\label{fig:reorder2}
		\centering{\includegraphics[scale=0.65]{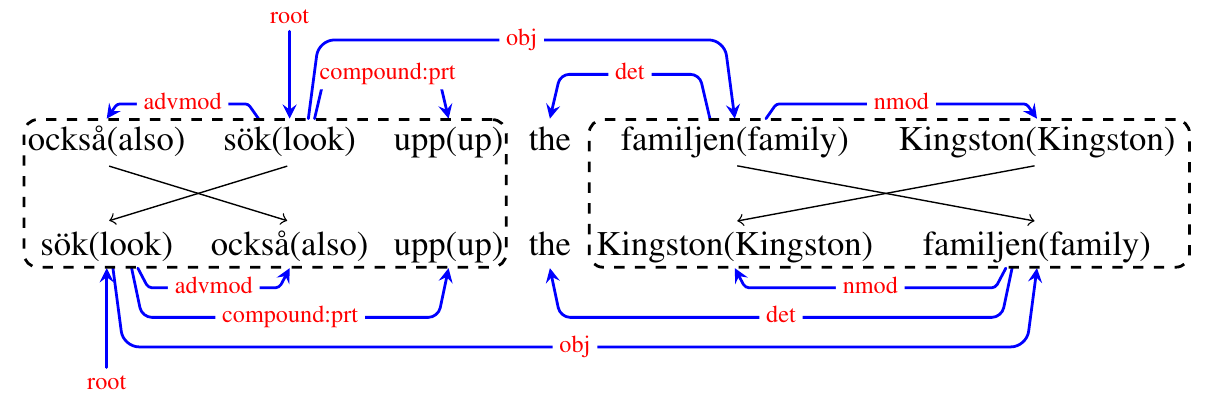}}
	}
\caption{Two examples of sentence reorder. }\label{reorder}
\end{center}
\end{figure}

\subsection{Sentence Reordering}
Continuous target spans are reordered to make the final code-mixed sentence
contain grammatical phrases in the target language.
Figure \ref{fig:reorder1} shows one example of full sentence reordering.
We can see that the word order by word-level substitutions on the source words
is different with the order of the machine-translated sentence.
Thus we adjust the leaf nodes, letting the word order strictly follow the machine-translated sentence order
For example,
the word ``vi'' is moved from the first position into the third position,
and similarly the word ``mu'' is moved from the last position into the first position.

Concretely, we perform the word reorder by the span level, extracting all the continuous spans of target words,
because the target language words may be interrupted by source language words.
Then we reorder the words in each span according to their order in the machine translation outputs.
Figure \ref{fig:reorder2} shows another example, where there are two spans separated by the English word ``the''.
Each span is reordered individually.
We do not consider the inconsistent orders inter the spans in this work.
Note that this step does not change any dependency arc between words.

\section{Experiments}
We conduct experiments to verify the effectiveness of our proposed models in this section.
\subsection{Settings}
Our experiments are conducted on the Google Universal Dependency Treebanks (v2.0) \cite{mcdonald2013universal,nivre2016universal},
using English as the source language,
and choosing six languages,
including Spanish (ES), German (DE), French (FR), Italian (IT), Portuguese (PT)
and Swedish (sv), as the target languages.
Google Translate\footnote{https://translate.google.com/ at Oct, 2018} is used to translate
the sentences in the English training set into other languages.
In order to generate high-quality word-level alignments,
we merge the translated sentence pairs and the parallel data of EuroParl \cite{koehn2005europarl}
to obtain word alignments.
We use the fastAlign tool \cite{dyer2013simple} to obtain word alignments.

We use the cross-lingual word embeddings and clusters by
\newcite{guo2016representation} for the baseline system.
The dimension size of word embeddings is 50 and
the word cluster number across of all languages is 256.

For network building and training,
we use the same setting as \newcite{dozat2016deep},
including the dimensional sizes,
the dropout ratio,
as well as the parameter optimization method.
We assume that no labeled corpus is available for the target language.
Thus training is performed for 50 iterations over the whole training data without early-stopping.

To evaluate dependency parsing performances,
we adopt UAS and LAS as the major metrics,
which indicate the accuracies of unlabeled dependencies
and labeled dependencies, respectively.
We ignore the punctuation words during evaluation following previous work.
We run each experiment 10 times and report the averaged results.

\subsection{Models}
We compare performances on the following models:
\begin{itemize}
  \item \texttt{Delex} \cite{mcdonald2013universal}: The delexicalized BiAffine model without cross-lingual word embeddings and clusters.
  \item \texttt{Src} \cite{guo2015cross}: The BiAffine model trained on the source English treebank only.
  \item \texttt{PartProj} \cite{lacroix2016frustratingly}: The BiAffine model trained on the corpus by projecting only the source dependencies
  involving high-confidence alignments into target sentences. Note that the baseline only draws the idea from \newcite{lacroix2016frustratingly},
  and the two models are significant different in fact.
  \item \texttt{Tgt} \cite{tiedemann2016synthetic}: The BiAffine model trained on the fully translated target treebank only.
  \item \texttt{Src+Tgt}: The BiAffine model trained on the combination dataset of the source and fully translated target treebanks.
  \item \texttt{Mix}: The BiAffine model trained on the code-mixed treebank only.
  \item \texttt{Src+Mix}: The BiAffine model trained on the combination dataset of the source and code-mixed treebanks.
\end{itemize}
The \texttt{Src} and \texttt{Tgt} methods have been discussed in Section 4.
The \texttt{PartProj} model is another way to leverage imperfect word alignments \cite{lacroix2016frustratingly}.
The training corpus of \texttt{PartProj} may be incomplete dependency trees with a number of words missing heads,
because no word is deleted from machine translation outputs.
The POS tags of words in \texttt{PartProj} with low-confidence alignments
are obtained by a supervised POS tagger \cite{yang2018design} trained on the corresponding universal treebank.

\subsection{Development Results}
We conduct several developmental experiments on the Swedish dataset
to examine important factors to our model.



\begin{figure}[t]
\begin{center}
\includegraphics[scale=0.95]{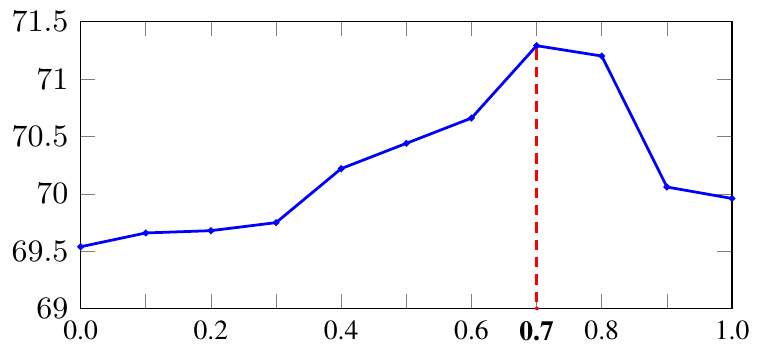}
\caption{The LAS relative to the translation ratio $\lambda$. }\label{dev-lambda}
\end{center}
\end{figure}

\setlength{\tabcolsep}{12pt}
\begin{table}[t]
\begin{center}
\begin{tabular}{l|rr}
\hline
Model &  UAS & LAS \\ \hline
\texttt{Src}   &   79.52   &  69.54   \\
\texttt{Tgt}  &   79.34   &  69.96  \\
\bf\texttt{Mix} &   \bf 80.33   &  \bf 71.29  \\ \hline
\texttt{Src + Tgt} &   80.12  &  71.16  \\
\bf\texttt{Src + Mix} &  \bf 80.91   &  \bf 71.73  \\
\hline
\end{tabular}
\caption{Experiments of corpus mixing.} \label{table:corpus:dev}
\end{center}
\end{table}

\setlength{\tabcolsep}{6pt}
\begin{table}[t]
\begin{center}
\begin{tabular}{l|rr}
\hline
Model &  UAS & LAS \\ \hline
\bf\texttt{Mix} &   \bf 80.33   &  \bf 71.29  \\ \hline
$-$Sentence Reordering &   79.79  &  70.47  \\
$-$Word Deletion &   79.82   &   70.64  \\ \hline
$-$Both &   79.46   &   69.59  \\
\hline
\end{tabular}
\caption{Ablation experiments.} \label{table:ablation}
\end{center}
\end{table}

\subsubsection{Influence of The Translation Ratio $\lambda$ }

Our model has an important hyper-parameter $\lambda$ to control the percentage of translation.
Figure \ref{dev-lambda} shows the influence of this factor,
where the percentages increase from 0 to 1 by intervals of 0.1.
A $\lambda$ of 0 gives our baseline by using the source treebank only.
As the $\lambda$ grows, more source words are translated into the target.
We can see that the performance improves after translating some source dependencies into the target,
demonstrating the effectiveness of syntactic transferring.
The performance reaches the peak when $\lambda = 0.7$,
but there is a significant drop when $\lambda$ grows from $0.8$ to $0.9$.
This can be because the newly added dependency arc projections are mostly noisy.
This sharp decrease indicates that noise
from low-confidence word alignments can have strong impact on the performance.
According to the results, we adopt $\lambda = 0.7$ for code-mixed treebanking.

\setlength{\tabcolsep}{2pt}
\begin{table*}[!ht]
\begin{center}
\begin{tabular}{c|cc|cc|cc|cc|cc|cc|cc}
\hline
\multirow{2}{*}{Lang.} &  \multicolumn{2}{c|}{ \texttt{Delex}} &  \multicolumn{2}{c|}{ \texttt{PartProj}} &
\multicolumn{2}{c|}{ \texttt{Src}}  &  \multicolumn{2}{c|}{ \texttt{Tgt}}
&  \multicolumn{2}{c|}{ \texttt{Src + Tgt} }  &  \multicolumn{2}{c|}{ \texttt{Mix}}  &  \multicolumn{2}{c}{\bf \texttt{Src + Mix}}  \\ \cline{2-15}
 &  UAS & \multicolumn{1}{c|}{LAS}  & UAS & \multicolumn{1}{c|}{LAS}  &  UAS & \multicolumn{1}{c|}{LAS} &  UAS & \multicolumn{1}{c|}{LAS} &
UAS & \multicolumn{1}{c|}{LAS} &  UAS & \multicolumn{1}{c|}{LAS} &  UAS & LAS \\ \hline
DE  & 64.10 & 53.77 & 69.90 & 61.28 & 66.87 & 57.46 & 70.84 & 62.30 & 72.41 & 63.74 & 71.41 & 63.46 & \bf 72.78 & \bf 64.38  \\
ES  & 71.53 & 63.33 & 75.81 & 66.83 & 75.63 & 65.85 & 76.49 & 67.39 & 77.00 & 67.95 & 81.18 & 71.80 & \bf 81.44 & \bf 71.66  \\
FR  & 75.13 & 67.26 & 75.54 & 67.63 & 78.13 & 70.63 & 76.91 & 69.39 & 78.75 & 71.17 & 83.20 & 76.32 & \bf 83.77 & \bf 76.48  \\
IT  & 77.71 & 69.27 & 77.71 & 69.27 & 81.11 & 72.83 & 79.30 & 71.65 & 81.56 & 74.09 & 85.30 & 77.43 & \bf 86.13 & \bf 78.38  \\
PT  & 74.03 & 67.70 & 79.44 & 71.30 & 77.37 & 69.36 & 78.32 & 70.67 & 79.73 & 71.84 & 83.54 & 75.34 & \bf 84.05 & \bf 75.89  \\ \hline
AVG & 72.50 & 64.27 & 75.68 & 67.26 & 75.82 & 67.23 & 76.37 & 68.28 & 77.89 & 69.76 & 80.93 & 72.87 & \bf 81.63 & \bf 73.36  \\
\hline
\end{tabular}
\caption{Final results.} \label{table:corpus:test}
\end{center}
\end{table*}

\subsubsection{Mixing with Source TreeBank}
We investigate the effectiveness of the source treebank by merging it into the translated treebanks.
First, we show the model performances of \texttt{Src}, \texttt{Tgt} and \texttt{Mix},
which are trained on the individual treebanks, respectively.
Then we merge the source treebank with the two translated treebanks,
and show the results trained on the merging corpora.
Table \ref{table:corpus:dev} shows the results.
According to the results, we can find that the source treebank is complementary with the translated treebanks.
Noticeably, although \texttt{Src + Mix} gives the best performance,
its improvement over \texttt{Mix} is relatively smaller than that of \texttt{Src + Tgt} over \texttt{Tgt}.
This is reasonable as the code-mixed treebank contains relatively more source treebank content
than the fully translated target treebank.

\subsubsection{Ablation Studies}
The overall translation is conducted by three steps as mentioned in Section 4,
where the first word substitution is compulsory,
and the remaining two steps aim to build better mixed dependency trees.
Here we conduct ablation studies to test the effectiveness of word deletion and sentence reordering.
Table \ref{table:ablation} shows the experimental results.
We can see both steps are important for dependency tree translation.
Without word deletion and sentence reordering,
the \texttt{mix} model shows decreases of 0.82 and 0.65 on LAS, respectively.
If both are removed,
the performance is only comparable with the baseline \texttt{src} model (see Table \ref{table:corpus:dev}).

\subsection{Final Results}

We show the final results of our proposed models in Table \ref{table:corpus:test}.
As shown,
the model \texttt{Tgt} gives better averaged performance compared to \texttt{Src}.
However, its results on French and Italian are slightly worse,
which indicates that noise from translation impacts the quality of the projected treebank.
The model \texttt{Mix} gives much better performance compared with \texttt{Src},
demonstrating the effectiveness of structural transfer.
\texttt{Mix} also outperforms \texttt{Tgt} significantly,
obtaining average increases of $80.93-76.37=4.56$ points on UAS and $72.87-68.28=4.59$ points on LAS, respectively.
The best setting is the model \texttt{Src + Mix}, trained on the combined corpus of the source and code-mixed treebanks,
which gives better performance than solely code-mixed treebanks.

By comparing with the delexicalized model \texttt{Delex},
we can see that lexicalized features are highly useful for cross-lingual transfer.
For the \texttt{PartProj} model,
we conduct preliminary experiments on Swedish to tune the ratio of the projected dependencies.
The results show that the difference is very small ($\delta = 0.24$ for UAS) between 0.9 to 1.0,
and the performance degrades significantly as the ratio decreases below 0.9.
The observation indicates that this method is probably not effective for filtering
low-confidence word alignments.
The final results confirm our hypothesis.
As shown in Table \ref{table:corpus:test}, the \texttt{PartProj} model gives only comparable performance with \texttt{Src}.
One possible reason may be the unremoved target words
(if the words are removed, the \texttt{PartProj} model with ratio 1.0 will be identical to \texttt{Tgt}),
which have been demonstrated noisy previously \cite{tiedemann2016synthetic}.

\setlength{\tabcolsep}{5pt}
\begin{table}[t]
\begin{center}
\begin{tabular}{c|ccccc}
\hline
Model & DE  &  ES & FR  &  IT   & PT  \\ \hline
\multicolumn{6}{c}{TreeBank Transferring} \\ \hline \hline
\texttt{This}             & 72.78 & \bf 81.44  & \bf 83.77  & \bf 86.13  & \bf 84.05   \\
\texttt{Guo15}            & 60.35 & 71.90  & 72.93   & -- & --  \\
\texttt{Guo16}            & 65.01 & 79.00  & 77.69   & 78.49 & 81.86  \\
\texttt{TA16}             & \bf 75.27 & 76.85 & 79.21 & -- & --   \\  \hline \hline
\multicolumn{6}{c}{ Annotation Projection } \\ \hline
\texttt{MX14}             & 74.30 & 75.53  & 70.14  & 77.74 & 76.65   \\
\texttt{RC15}              & \bf 79.68  & \bf 80.86  & \bf 82.72 & \bf 83.67 &  \bf 82.07    \\ 
\texttt{LA16}    & 75.99  & 78.94 & 80.80  & 79.39  & --   \\ \hline \hline
\multicolumn{6}{c}{TreeBank Transferring + Annotation Projection} \\ \hline
\bf \texttt{RC17}    & \bf 82.1  & \bf 82.6 & \bf 83.9  & \bf 84.4  & \bf 84.6   \\ 
\hline
\end{tabular}
\caption{Comparison with previous work (UAS).} \label{table:corpus:compare}
\end{center}
\end{table}

\subsection{Comparison with Previous Work}
We compare our method with previous work in the literature.
Table \ref{table:corpus:compare} shows the results,
where the UAS values are reported.
Our model denoted by \textbf{\texttt{This}} refers to the model of \texttt{Src + Mix}.
Note that these models are not directly comparable due to the setting and baseline parser differences.
The first block shows several models by directly transferring gold-standard source treebank knowledge into the target side,
including the models of \texttt{Guo15} \cite{guo2015cross},
\texttt{Guo16} \cite{guo2016representation} and \texttt{TA16} \cite{tiedemann2016synthetic}.
Our model gives the best performance with one exception on the German language.
One possible reason may be that \texttt{TA16} has exploited multiple sources of treebanks besides English.

The second block shows representative annotation projection models,
including \texttt{MX14} \cite{ma2014unsupervised}, \texttt{RC15} \cite{rasooli2015density}, \texttt{LA16}.
The models of annotation projection can be complementary with our work,
since they build target training corpus from raw parallel texts.
The best-performed results of the \texttt{RC17} model \cite{rasooli2017cross} have demonstrated this point,
which can be regarded as a combination of the dictionary-based treebank translation\footnote{The method has been demonstrated worse than \texttt{TA16} in \newcite{tiedemann2016synthetic}.} \cite{zhao2009cross} and \texttt{RC15}.

\subsection{Analysis}
We conduct experimental analysis on the Spanish (ES) dataset to show the differences
between the models of \texttt{Src}, \texttt{Tgt} and \texttt{Mix}.

\begin{figure}[t]
\begin{center}
\includegraphics[scale=1.0]{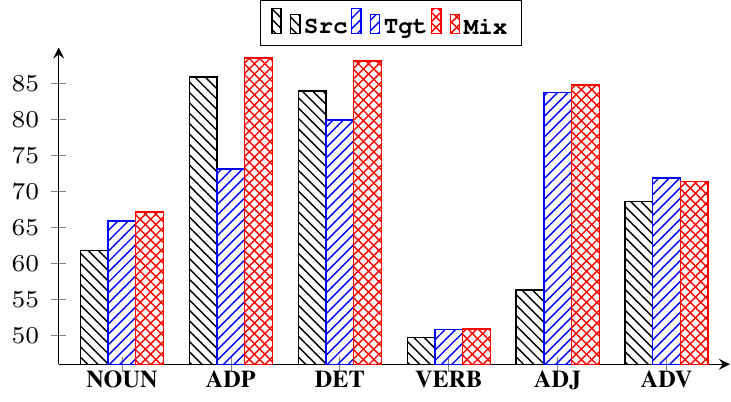}
\caption{Performance relative to POS tags (F-score). }\label{analysis:tag}
\end{center}
\end{figure}

\subsubsection{Performance Relative to POS Tags}
Figure \ref{analysis:tag} show the F-scores of labeled dependencies on different POS tags.
We list the six most representative POS tags.
The \texttt{Mix} model achieves the best F-scores on 5 of 6 POS tags,
with the only exception on tag ADV which has no significant difference with the \texttt{Tgt} model.
The \texttt{Mix} and \texttt{Tgt} models are much better than the \texttt{Src} model as a whole,
especially on the POS tag ADJ where an increase of over 20\% has been achieved.
In addition, we find that the \texttt{Src} model
can significantly outperform the \texttt{Tgt} model on ADP and DET.
For Spanish, ADP words are typically ``de'', ``en'',  ``con'' and etc.,
which behave similarly to the English words such as ``'s'', ``to'' and ``of''.
The Spanish words of DET include ``el'', ``la'',  ``su'' and etc.,
which are similar to the English words such as ``the'' and  ``a''.
These words are highly ambiguous for automatic word alignment.
The results indicate that our \texttt{Mix} model can better handle these word alignment noise,
mitigating their negative influence of treebank translation,
while the \texttt{Tgt} model suffers from such noise.

\begin{figure}[t]
\begin{center}
\includegraphics[scale=1.0]{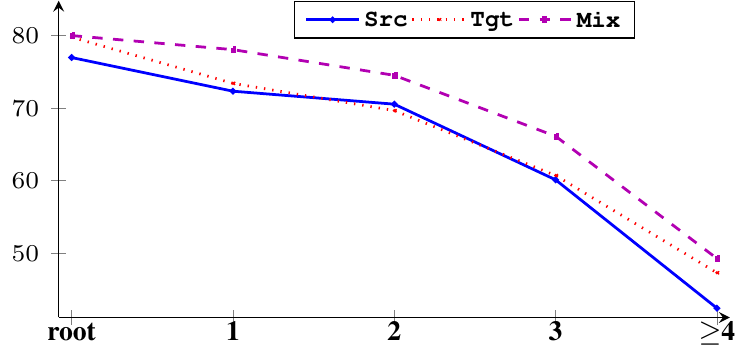}
\caption{Performance relative to arc distances (F-score). }\label{analysis:dist}
\end{center}
\end{figure}

\subsubsection{Performance Relative to Arc Distances}
Figure \ref{analysis:dist} show the F-scores of labeled dependencies by different arc distances.
Particularly we treat the root type as one special case.
According to the results, the \texttt{Mix} model performs the best over all distances,
indicating its effectiveness on treebank transferring.
The \texttt{Tgt} model achieves better performance than the \texttt{Src} model with one exception on distance 2.
We look into the dependency patterns of distance 2 arcs further,
finding that the dependency arc $\textbf{ADP}^{\overset{}{\bm{\curvearrowleft}}}\textbf{*}$ accounts for over 30\%,
and it is the major source of errors.
The finding is consistent with that on POS tags,
denoting the effectiveness of the code-mixed treebank in handling noise.
In addition, as the distance increases the performance drops gradually.
The F-score of root dependency is the highest.

\section{Conclusion}
We proposed a new treebank translation method for unsupervised cross-lingual dependency parsing.
Unlike previous work, which adopts full-scale translation for source dependency trees,
we investigated partial translation instead, producing synthetic code-mixed treebanks.
The method can better leverage imperfect word alignments between source and target sentence pairs,
translating only high-confidence source sentential words,
thus generating dependencies in high-quality.
Experimental results on Universal Dependency Treebak v2.0
showed that partial translation is highly effective,
and code-mixed treebanks can give significantly better results than full-scale translation.

Our method is complementary with several other methods for cross-lingual transfer,
such as annotation projection,
and thus can be further integrated with these methods.

\section*{Acknowledgments}
We thank all anonymous reviewers for their valuable comments.
Several suggestions are not integrated in this version, i.e., more experiments on really low-resource languages,
and detailed analysis on more languages.
We will supplement later on the webpages of the authors.
This work is supported by National Natural Science Foundation of China (NSFC) grants 61602160, U1836222 and 61672211.

\bibliography{reference}
\bibliographystyle{acl_natbib}

\end{document}